# Speeding Up Action Recognition Using Dynamic Accumulation of Residuals in Compressed Domain

Ali Abdari
Department of Engineering
Faculty of Electrical and Computer Engineering
Kharazmi University
Tehran, Iran
Email: std_ali.abdari@khu.ac.ir

Pouria Amirjan
Department of Engineering
Faculty of Electrical and Computer Engineering
Kharazmi University
Tehran, Iran
Email: std_pamirjan@khu.ac.ir

Azadeh Mansouri
Department of Engineering
Faculty of Electrical and Computer Engineering
Kharazmi University
Tehran, Iran
Email: a_mansouri@khu.ac.ir

**Abstract**

With the widespread use of installed cameras, video-based monitoring approaches have seized considerable attention for different purposes like assisted living. Temporal redundancy and the sheer size of raw videos are the two most common problematic issues related to video processing algorithms. Most of the existing methods mainly focused on increasing accuracy by exploring consecutive frames, which is laborious and cannot be considered for real-time applications. Since videos are mostly stored and transmitted in compressed format, these kinds of videos are available on many devices. Compressed videos contain a multitude of beneficial information, such as motion vectors and quantized coefficients. Proper use of this available information can greatly improve the video understanding methods' performance. This paper presents an approach for using residual data, available in compressed videos directly, which can be obtained by a light partially decoding procedure. In addition, a method for accumulating similar residuals is proposed, which dramatically reduces the number of processed frames for action recognition. Applying neural networks exclusively for accumulated residuals in the compressed domain accelerates performance, while the classification results are highly competitive with raw video approaches.

# 1 Introduction

The increasing need for video-based applications reveals the importance of video understanding tasks. Video is massive data and has high redundancy. The majority of internet traffic is attributed to video transmissions due to the proliferation of video applications such as those used by healthcare facilities, smart cars, and real-time communications.

For real-time applications, it is essential to use methods with low complexity for fast and online video processing. Although many recently presented methods for action recognition try to boost accuracy, most of them ignore the high complexity of processes, making them improper for real-time applications. These methods often use pixel domain features like optical flows, which are too complex to be practical to use in online applications.

Nowadays, compressed videos are available on numerous electronic devices. Compressed videos contain useful and available data, including motion vectors and quantized coefficients. Exploiting this accessible information appropriately would obviate the need to extract time-consuming features like optical flows. CCTV cameras mainly use video compression standards for storing or transmitting data. These cameras surround many places, such as equipped houses, hospitals, stores, and other public places. Thus, it is of practical interest to extract the information directly from the compressed video signal.

Practically, reliable smart systems should have characteristics such as efficiency, accuracy, and responsiveness. Efficiency in intelligent systems can be measured by energy and storage consumption. Quick extraction of appropriate features occupies center stage in fast video action detection and recognition. As a result, when compressed videos are available, using compressed domain information can improve efficiency. In fact, by partially decoding a compressed video, we can obtain the residuals containing important spatio-temporal information.

Analyzing surveillance videos are required for many purposes. Since these kinds of videos contain numerous or infinite frames that are highly similar to their adjacent frames, a method that processes all of the consecutive frames would be inefficient. In our early works, we presented a framework, which utilizes compressed domain residuals for action recognition explained in [1, 2]. This paper exploits the residual as the compressed domain data and analyzes the results of replacing the original RGB frames in the raw domain with the compressed domain data. Also, a dynamic accumulation mechanism is proposed to build a powerful augmented residual frame. The achieved signal contains robust and long-term information. Thus, at the same time that we process much fewer frames, we also obtain acceptable efficiency as well.

## 2 Related works

The topic of action recognition in video is challenging and can be broken down into two main categories. The following paragraphs explain how some methods use raw frame information, such as appearances and motions, while others use compressed domain accessible information.

### 2.1 Raw frame features

As events in a sequence of frames occur during time and space, detecting spatio-temporal features is indispensable in action recognition. Histogram of Gradients (HOG) [3] is one of the approaches representing local changes in a frame. Some video information like motions exists along consecutive frames compared to local information available in each frame independently. Since optical flow contains motion information along consecutive frames, handcrafted optical flow-based features are considered one group of the most popular methods[4, 5]. Furthermore, motion boundary histogram (MBH), which extracts features employing the gradient of optical flows, is presented in [6]. In [7], the motion boundary of optical flows for a dense sampling of interest points is utilized. An enhanced KLT tracker tracks interest points over time and accumulates in a three-dimensional trajectory structure. In this case, multiscale descriptors are employed for representation. In [8], the authors proposed a multiple-subsequence-combination method, which divides the video into several consecutive subsequences. Researchers in [9] proposed a method to employ salient proto-objects for unsupervised discovery of object and object-part candidates using HOF.

Convolutional neural networks (CNNs) are used frequently in different approaches to analyzing images. In [5], some pooling methods are employed on feature vectors extracted by CNN in order to provide better temporal representation. In [10, 11], two-stream networks are used to make use of both temporal and spatial features. Optical flows are fed to the temporal network, and raw decoded RGB pictures are fed to the spatial network. Additionally, in another research [12], kernelized rank pooling is proposed to represent actions during consecutive frames.

Considering several consecutive frames instead of a single frame could increase the accuracy of the approach since we can obtain valuable spatial and temporal features. In some research like [13], 3D-CNNs are used and fed by a bunch of frames. Moreover, in order to aggregate the frame-wise features, RNN and LSTM areas are used in some research to manage the time-series information of video [14, 15]; however, utilizing these types of networks is too time-consuming.

In [16] another type of video, first-person datasets, are explored. Due to camera motion, action recognition in first-person datasets could be more difficult. The authors proposed three-stream architecture designed to extract appearance, motion, and camera motion features by using convolutional networks. After extracting feature maps, maximum and average values are considered for the fusion step to obtain a unique feature map. In [17], the same architecture with a new correlation-based fusion approach is utilized. In these two articles, for the classification step, an LSTM network has been exploited.

### 2.2 Compressed domain features

Using compressed domain elements has become more prevalent in different computer vision fields like object detection [18], object tracking [19, 20], and segmentation [21].

Similar to the raw frame approaches, handcrafted and deep learning methods are two strategies for analyzing compressed videos. Different kinds of motion estimation methods, including Full Search (FS), large Diamond Search (DS), and small Diamond Search (sDS) flow for fast activity recognition, are presented in [22]. DCT coefficients contain structural information that can be extracted through these available elements. In [23], handcrafted features are explored using DCT coefficients, and residual histograms are formed as video representations. Although motion vectors are block-based and are not as accurate as optical flows, calculating optical flows is too complex compared to motion vectors accessible via compressed video. In [24], a method is proposed using the benefit of optical flow and motion vectors simultaneously. First, optical flows are applied to train a network, and motion vectors are employed to fine-tune it. Then in the test step, by using motion vectors as the input of the trained network, the method prevents calculating optical flows in run time. Likewise, in [25], a Discriminative Motion Cue network(DMC-Net) has been proposed. A DMC generator is utilized, receiving motion vectors and residuals as input. This generator improves coarse motion vectors to be more accurate by employing optical flow in offline time to train a generative model.

Consecutive raw frames are far similar, and most of their differences are associated with motion information. Some research [26, 27] utilize the difference between frames containing the motion information instead of the raw frames in their processes. The residual information in compressed videos can be obtained by applying Inverse DCT (IDCT) on available DCT coefficients in a compressed video file. Since the residuals are very close to the difference between consequent frames, it would be good to use residuals instead of the difference of frames as the input of CNN. The less redundant information is given as input to the network, the more accurate output will be obtained [28]. The raw and residual frames are fed into the CNN to better illustrate the residual information's importance. The heatmap of the network is extracted and illustrated in Figure 1. It is clearly shown that when we use residual information instead of raw frames, the most relevant points are better illustrated in a heatmap image. In other words, it means that CNN can extract better features by employing residual information. In fact, the residual component as a spatio-temporal information played an important role and RGB frames can be replaced by residual information properly.

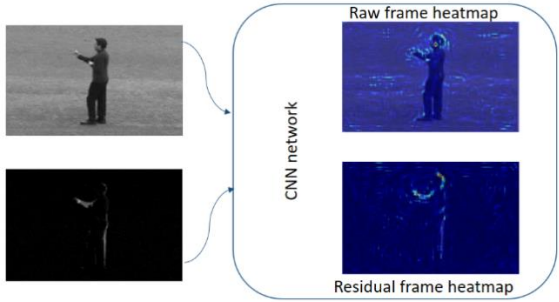

**Figure 1.** Comparison of heatmaps between the residual and raw frame

In a recent research [29], a teacher-student framework was employed, in which the teacher network is a pre-trained network using raw frames, and the student network is fed by I-frames information and accumulated P-frames residuals (i.e., for each GOPs) as input and some information transferred from raw domain network.

In this paper, firstly, we scrutiny the approach of utilizing residual information in action recognition and analyze the results of replacing the original RGB frames in the raw domain with the compressed domain data. Then, a dynamic accumulation algorithm is proposed to construct robust and long-term spatiotemporal frames. This technique provides greater efficiency and decreases the number of frames dramatically.

The rest of the paper's structure is as follows: Section **3** explains the proposed method. Experimental results are presented in section **4**, and finally, the conclusion is discussed in section **5**.

## 3 Proposed Method

### 3.1 Employing Residual Information

Compressed videos already carry beneficial and available information, such as motion vectors and quantized coefficients. During the compression process of a video, frames will be divided into some macroblocks, and for each macroblock, the best match will be found by searching in a reference frame(s). Then the difference between every macroblock and the related best match block will be calculated. The following statement illustrates this difference for each pixel:

$$R_i = M_i - M_i^P \quad (1)$$

In which $R_i$, $M_i$ and $M_i^P$ show the residual block, the original and the predicted best match macroblocks, respectively. The quantization of discrete cosine transform of differences will be encoded and stored in compressed video format. $R_q$ illustrates the result of applying $DCT$ and quantization ($Q$) on residual information as follows:

$$R_q = Q(DCT(R)) \quad (2)$$

So, $Rq$ and a 2-D motion vector related to the macroblock are generated and transmitted for every block.

The size of the macroblocks depends on compression standards. In some compression methods, the size of the macroblocks is fixed, while in the recent standards, it is feasible to slice the frames by different sizes. Smaller macroblocks mostly encode regions with fast motions and complex structures, and larger ones encode regions with simple structures. In the MPEG2 standard, the size of macroblocks is fixed on 16×16.

We assume that a compressed video file is available in the MPEG.2 format in the proposed method implementation. Then the compressed video is partially decoded to obtain residues. In this case, we should apply the dequantization ($Q^{-1}$) process followed by the inverse discrete cosine transform ($IDCT$) to estimate the residual information in the spatial domain.

$$\tilde{R} = IDCT(Q^{-1}(R_q)) \quad (3)$$

We use the obtained residual as the input of a pre-trained neural network for feature extraction. In this paper, *ImageNet VGG-f* is considered as the pre-trained model, which is trained using *MatConvNet2* on the $ImageNet$ dataset.

The model's input is the residual estimation obtained by partially decoding the compressed videos. These inputs are resized into $224 \times 224$. Moreover, the normalization step is considered based on a calculated average image specified in the pre-trained network. Outputs of the $18^{th}$ layer of size $4096$ are considered as the extracted feature. Every input produces a $4096 \times 1$ feature vector. As a result, an input video of length $M$ will generate a matrix of size $4096 \times M$.

SVM classifier with χ2 kernel for action classification is used for the classification step. The size of the input in each classifier is unique, but the output of the feature extraction step has a size $4096 \times M$. We use the Pooled Time Series approach to reshape the matrix to a unique size [5].

We conducted a temporal partitioning method for data augmentation and action segmentation. Firstly, the temporal partitioning of motion information is presented using optical flow images in [30]. In this paper, the residual information formed the input videos. Each residual video is then decomposed into several temporal partitions, and a max-pooling operator is applied to the partitions. The action of each partition is recognized, and finally, the class of the input video is obtained using a voting strategy. For the implementation, we consider the partition size eight; hence, the main video is decomposed into eight segments. The pooling method is applied to each part and produces eight decisions. To classify the action of the input video, we vote among these decisions.

Generally, the authors propose the main idea of using residual information in [2]. In fact, we showed that the residuals carry enough information to replace the original RGB frames. We investigate the possibility of using this low complexity approach for healthcare monitoring as a case of real-time applications in [1].

In the next section, we try accumulating the residual information to magnify the relative information. The accumulation strategy is used in [28, 29] in the fixed time step, but in the following section, we introduce the dynamic and content-based accumulation approach and explain the effectiveness of the proposed method in experimental results.

### 3.2 Accumulating Residual Frames

In many cases, residuals contain few values, especially when the movement within the video is too slow. Also, in some types of videos, like surveillance ones, the number of frames is too high, making it inapplicable to analyze each frame individually. Feeding a single residual to CNN in these cases could impinge upon time efficiency, resulting in a dramatic reduction in performance. This section proposes a dynamic accumulation algorithm to accumulate similar residuals and provide the accumulated frame for CNN.

To better illustrate the efficiency of the accumulated residuals, Figure 2 shows the accumulated residuals and a single residual frame, which represents a person moving his hand. The magnified parts illustrate how accumulated residual frames can represent the motion more clearly than a single residual.

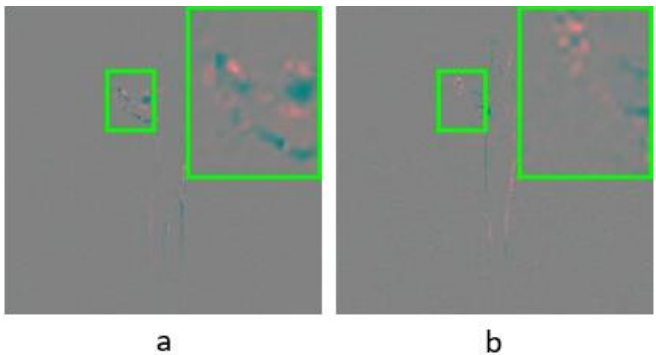

**Figure 2.** Magnified images of a) accumulated residuals and b) single residual frame

In this method, feeding CNN by residuals has improved by deploying a temporal window, which moves over residual frames. This window contains the similarity values between consequent residuals.

*Similarity* measures the amount of similarity between two residual frames. The higher value of this measure means the two residual frames are more similar. The range of the value varies between zero and one. The closer this criterion is to one, the more similar the two residuals are. Statement (4) illustrates the *similarity* formula, in which $R_i$ is the current residual and $R_{i-1}$ is the previous one. Also, $c$ is a constant value.

$$SIMILARITY_i = \frac{2 \times R_i \times R_{i-1} + c}{R_i^2 + R_{i-1}^2 + c} \quad (4)$$

There is a window size demonstrating how many consecutive residuals information should be stored in the history of the window. If we assume the window size is *N*, the window contains the history of the last *N* consecutive residual frames *similarity*. During the residual frames' processing for each new residual, the similarity of this residual with the previous one is calculated. If the similarity is more than the mean of the temporal window, we accumulate the current residual with the last group. On the other hand, if the similarity is less than the mean of the window, we assume we have entered a new motion event, then the previous group of residuals will be cut, and a new group containing the mentioned residual will be considered. The previous group members will be accumulated to be prepared as the CNN input.

Besides, the window members will be updated, and in each step, the window contains *N* last newest residuals similarity that *N* represents the window size. The steps of the algorithm are explained in more detail in Algorithm 1.

1. Add *N* (size of the Temporal Window) residuals similarities to the window and compute the mean of the vector containing residuals' similarity.
2. Calculate the next residual similarity with the last window member.
3. If the similarity is more than the mean of the window, this residual will be accumulated with the previous frames and go to step 5; otherwise, go to step 4.
4. The last group of accumulated residuals will be fed to the CNN; go to step 1.
5. Move forward the temporal window.
6. Update the mean of the window.
7. If there is any residual, go to step 2.

**Algorithm 1.** The steps of detecting and accumulating similar residuals.

By applying this dynamic accumulation algorithm, we improve the efficiency of the frame-by-frame approach. We feed a fewer number of inputs to the CNN for extracting features. Experimental results show that utilizing this method has decreased the processed residual frames by about 45%. An overview of the dynamic accumulation approach is represented in Figure 3.

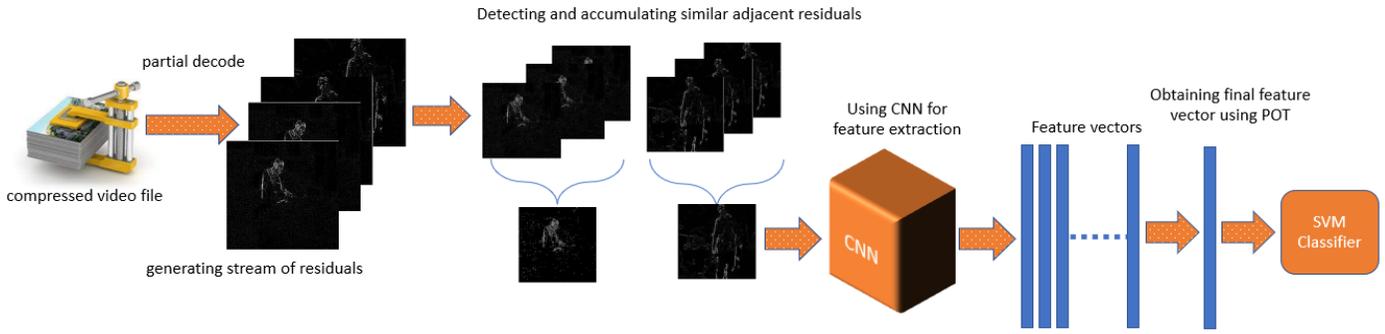

**Figure 3.** Overview of the dynamic accumulation approach. First, we partially decode compressed video to obtain a residual stream. By using the similarity formula and temporal window method, we detect adjacent residuals and accumulate them. After passing this step, strongly built residuals will be fed into the neural network to extract feature vectors. After employing POT [5], the final feature vector can be used for classification.

## 4 Experimental Results

The experimental results are divided into two test scenarios. In our first experiment, we used residual instead of RGB frames to identify the effectiveness of employing this information, and then we evaluated the proposed dynamic accumulation approach.

### 4.1 NUSFPID and JPL datasets

To evaluate the proposed method in first-person action recognition datasets, we test and compare the results using two first-person datasets: the NUSFPID [31] and JPL [32] datasets. The NUSFPID dataset contains eight interaction classes, including cell, door, pass, shake, throw, type, wave, and write, while JPL contains seven classes, including handshake, pet, hug, wave, point-converse, punch, and throw.

The results of the NUSFPID and JPL datasets are presented in Table 1, where "Residuals" demonstrate the proposed method, and the "DOF" and "RGB_F" represent "difference of RGB frames" and "RGB frames," respectively. Obtained results are compared to uncompressed domain methods. LRCN [15] stands for Long-term Recurrent Convolutional Network, whose results are represented in [27].

Also, the proposed framework is evaluated using the first-person JPL dataset. TSCF, TSDF, and KRP represent Three-stream Correlation Fusion, Three-stream Deep Fusion, and Kernelized Ranked Pooling, respectively. The presented results of [12, 16] are reported from [17]. It should be noted that all of the compared methods utilize raw RGB frames. Also, in three stream scenarios, time-consuming calculations of optical flows are necessary as well.

Comparing the proposed compressed domain method in Table 1 illustrates the acceptable performance since the residual represents the difference between the processed frames and their references. This information is usually well-aligned with the boundary of moving objects, which provides useful data for action recognition. As a result, our low complexity method utilizing available syntactic elements of the compressed domain can be employed for real-time applications.

**Table 1.** The overall accuracy of the proposed method and other methods on the NUSFPID and JPL datasets.

| methods | NUSFPID | JPL |
|---|---|---|
| LRCN(RGB_F) [15] | 68.90% | 59.50% |
| LRCN(DOF) [15] | 69.10% | 89.00% |
| LSTM(RGB_F) [27] | 69.40% | 70.00% |
| LSTM(DOF) [27] | 70.00% | 91.00% |
| TSDF(RGB_F) [16] | | 81.10% |
| TSDF(DOF) [16] | | 86.40% |
| KRP(RGB_F) [12] | | 73.80% |
| KRP(DOF) [12] | | 85.70% |
| TSCF(RGB_F) [17] | | 88.00% |
| TSCF(DOF) [17] | | 94.40% |
| Residuals(proposed) | 69.33% | 88.60% |

### 4.2 KIT datasets

We chose the KIT dataset to evaluate our dynamic accumulation method [33]. KIT is a daily living activity dataset recorded in the kitchen, containing five categories recorded from different sights of the kitchen. The proposed method has been evaluated in KIT_door and KIT_corner. Video frame sizes are reduced to 240×320 to speed up processing. In the following results, half of each class's videos were dedicated to the training step, and the others were used for the test step. In the following tables and figures, "No Acc" and "WS" denote without accumulation results and window size in the dynamic accumulation proposed method, respectively. KIT_door contains ten classes which are recorded from the door point of view. The results are given in Table 2.

TABLE 2. Per class results of the KIT_door dataset

| Actions | No Acc[2] | WS = 10 | WS = 30 | WS = 50 | HOGHOF [7] | [8] | proto-object + HOF [9] |
|---|---|---|---|---|---|---|---|
| Clear table | **86.39%** | 81.57% | 79.68% | 85.28% | **100.00%** | **100.00%** | 94.40% |
| Drink Coffee | 86.98% | **89.37%** | 81.80% | 84.81% | **100.00%** | **100.00%** | 91.20% |
| Cut vegetables | 67.44% | **72.38%** | 67.73% | 68.36% | **83.30%** | 71.00% | 65.20% |
| Empty dishwasher | **94.95%** | 94.68% | 92.19% | 92.76% | **100.00%** | **100.00%** | 98.80% |
| Peel vegetables | 79.97% | **82.33%** | 79.44% | 79.41% | **88.20%** | 57.00% | 72.60% |
| Eat pizza | **68.06%** | 64.55% | 63.06% | 59.58% | 78.60% | 86.00% | **88.30%** |
| Set table | **94.06%** | 90.66% | 85.44% | 93.02% | 92.80% | **100.00%** | 98.60% |
| Eat Soup | **90.28%** | 90.00% | 81.29% | 83.41% | **92.80%** | 71.00% | 88.20% |
| Sweep the floor | **100.00%** | 99.55% | 99.66% | 99.78% | **100.00%** | **100.00%** | 88.70% |
| Wipe table | **98.69%** | 97.70% | 96.13% | 96.42% | **100.00%** | **100.00%** | 93.40% |

KIT_corner contains seven classes, whose results are illustrated in Table 3. As shown in Table 2, despite the lower computational costs of the proposed compressed domain method using dynamic accumulation, the obtained results are more accurate for some classes in window size ten. Generally, the per-class results are better for the KIT_corner dataset in the dynamic accumulation scenario. Using the proposed dynamic accumulation method can reduce the redundancy of data and provide meaningful information for the network as an input, resulting in a better performance for real-time applications.

TABLE 3. Per class results of the KIT_corner dataset

| Class Name | No Acc[2] | WS = 10 | WS = 30 | WS = 50 |
|---|---|---|---|---|
| Cut | 84.63% | **86.97**% | 85.18% | 84.32% |
| Dry | 96.13% | 97.24% | 95.54% | **97.35%** |
| Fry | **88.74%** | 86.62% | 87.14% | 84.23% |
| Peel | 85.54% | 85.16% | **88.28%** | 83.86% |
| Stir | 97.35% | 96.18% | 96.22% | **97.86%** |
| Wash | 99.77% | **100**% | 98.56% | **100**% |
| Wipe | 97.98% | 99.65% | **99.88%** | 98.87% |

Figure 4 and Table 4 clearly illustrate the comparison between the entire number of processed frames and the obtained accuracies, respectively, in four scenarios; without accumulation, window size 10, 30, and 50.

TABLE 4. Comparison of accuracies on KIT_door and KIT_corner datasets

| Dataset | No Acc[2] | WS = 10 | WS = 30 | WS = 50 |
|---|---|---|---|---|
| KIT_corner | 92.06% | 93.65% | 93.65% | 92.06% |
| KIT_door | 86.02% | 86.02% | 82.79% | 82.79% |

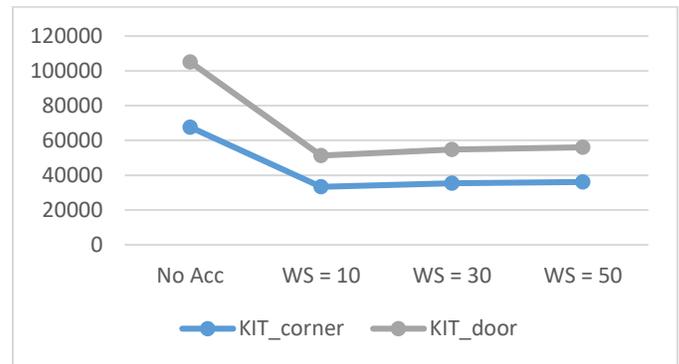

Figure 4. The number of processed frames for two KIT_door and KIT_corner datasets.

The number of processed residual frames in each dataset decreased by about 45% when we used the proposed accumulation approach. As seen in Table 4, using the accumulation method can have highly competitive accuracies while it processes much fewer frames.

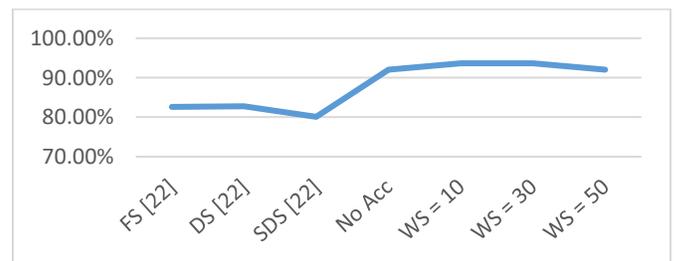

Figure 5. Comparison of the proposed method with other compressed domain approaches on the KIT_corner dataset.

Finally, the proposed method's performance is compared with the efficient motion estimation-based approach [22], which is obtained using three motion estimation scenarios. As illustrated in Figure 5, the proposed dynamic accumulated residuals can outperform other compressed domain approaches.

# 5 Conclusion

Compressed domain elements such as residuals contain important spatio-temporal information. We proposed a method for dynamically accumulating similar adjacent residuals to achieve robust and representative longer-term features. Thus, the network receives meaningful input, and redundancy is reduced. Consequently, the proposed method provides acceptable accuracies while using less processing power, making it suitable for real-time applications. When residuals are accumulated, and computation is reduced, the network becomes more efficient, which in turn keeps accuracy acceptable as well. Considering that the temporal window size has been fixed in this paper, our future work will analyze the effects of temporal window size and suggest some criteria to determine content-based temporal window size.